\documentclass[conf]{new-aiaa}
\usepackage[utf8]{inputenc}

\usepackage{graphicx}
\usepackage{amsmath}
\usepackage[version=4]{mhchem}
\usepackage{siunitx}
\usepackage{float}
\usepackage{graphicx}
\usepackage{subcaption}
\usepackage{longtable,tabularx}
\usepackage{booktabs}
\title{From Ground to Air: Noise Robustness in Vision Transformers and CNNs for Event-Based Vehicle Classification with Potential UAV Applications}

\author{Nouf Almesafri\footnote{Student, Applied Artificial Intelligence, Cranfield University.} \footnote{Senior Associate Researcher, Propulsion and Space Research Center, Abu Dhabi}}
\affil{Cranfield University, MK43 0AL Cranfield, UK}
\affil{Technology Innovation Institute, Masdar City, 9639 Abu Dhabi, UAE}
\author{Hector Figueiredo\footnote{Head of Technologies, UAS Business, Qinetiq}}
\affil{Qinetiq, MK43 7TA Bedford, UK}
\author{Miguel Arana-Catania\footnote{Senior Research Software Engineer, Digital Scholarship at Oxford, University of Oxford}}
\affil{University of Oxford, OX1 3BG Oxford, UK}

\begin{document}

\maketitle

\begin{abstract}
This study investigates the performance of the two most relevant computer vision deep learning architectures, Convolutional Neural Network and Vision Transformer, for event-based cameras. These cameras capture scene changes, unlike traditional frame-based cameras with capture static images, and are particularly suited for dynamic environments such as UAVs and autonomous vehicles. The deep learning models studied in this work are ResNet34 and ViT B16, fine-tuned on the GEN1 event-based dataset. The research evaluates and compares these models under both standard conditions and in the presence of simulated noise. Initial evaluations on the clean GEN1 dataset reveal that ResNet34 and ViT B16 achieve accuracies of 88\% and 86\%, respectively, with ResNet34 showing a slight advantage in classification accuracy. However, the ViT B16 model demonstrates notable robustness, particularly given its pre-training on a smaller dataset. Although this study focuses on ground-based vehicle classification, the methodologies and findings hold significant promise for adaptation to UAV contexts, including aerial object classification and event-based vision systems for aviation-related tasks.
\end{abstract}

\section{Introduction}
\lettrine{E}{vent-based} cameras, inspired by the human retina, represent a significant step forward in vision technology by capturing changes in brightness asynchronously \cite{posch2014retinomorphic}. Unlike traditional frame-based cameras, which capture static images at regular intervals, event-based cameras focus only on scene changes. This approach reduces redundancy, enables high temporal resolution, lowers power consumption, and minimizes motion blur \cite{posch2014retinomorphic}. These characteristics make event-based cameras particularly suitable for dynamic environments, such as UAVs \cite{paulo2023collision} and autonomous vehicles \cite{janai2020computer}. However, despite their potential, the effective integration of event-based data with deep learning models remains an underexplored area.\\

\begin{figure}[htp!]
    \centering
    \includegraphics[width=0.5\linewidth]{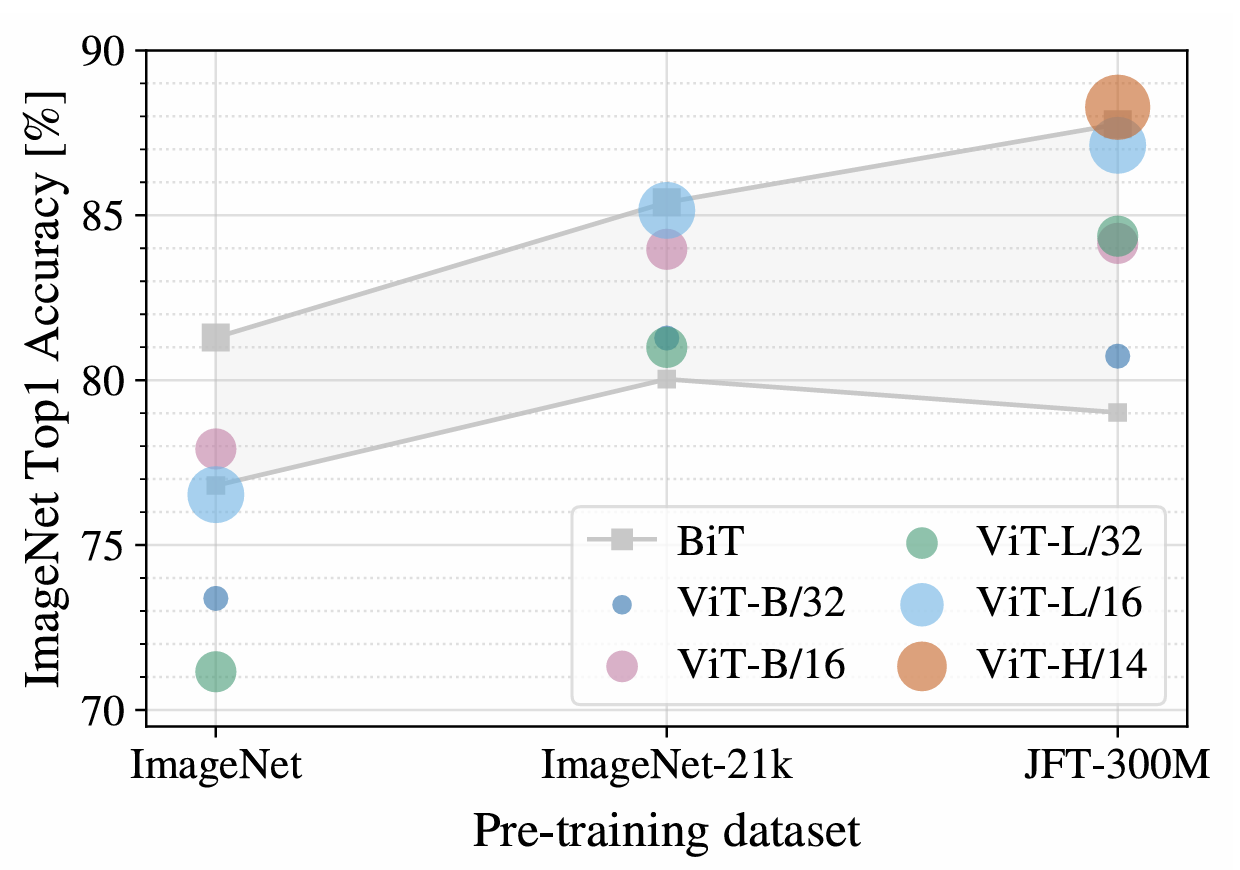}
    \caption{Image classification accuracy on datasets of increasing size. Large ViT models underperform compared to BiT ResNets (shaded area) when pre-trained on smaller datasets \cite{dosovitskiy2020image}.}
    \label{fig:pretrained_comb_figure}
\end{figure}

This study investigates the performance of two widely used deep learning models, ResNet34 \cite{he2016deep} and Vision Transformer (ViT) B16 \cite{dosovitskiy2020image}, in classifying objects using the GEN1 event-based dataset \cite{de2020large}. ResNet34, a well-established Convolutional Neural Network (CNN), serves as a baseline for comparison. On the other hand, ViT B16 represents a newer class of models based on the transformer architecture, which has recently gained attention for its success in various computer vision tasks \cite{dosovitskiy2020image}.  Figure \ref{fig:pretrained_comb_figure} shows a performance comparison between these different architectures on image classification tasks when the models are pre-trained on datasets of different sizes. The research explores how these models perform in both clean and noisy conditions, with a particular focus on noise types such as event shifts, polarity reversals, and event loss, which are commonly encountered in real-world data \cite{ben2022discussion}.\\

The primary aim of this study is to evaluate the robustness of ViT B16 compared to ResNet34 under noisy conditions. The research also seeks to determine whether ViT can offer an advantage in terms of accuracy and generalization, especially when the dataset size is limited. While the GEN1 dataset is tailored for automotive applications, the methodologies and findings have the potential to extend to other domains, including UAV-based imaging and autonomous vehicles.\\

Event-based cameras and deep learning models, such as those studied here, could play a pivotal role in UAV applications. For instance, event-based cameras' low latency and high temporal resolution make them ideal for real-time obstacle avoidance in dynamic environments, enabling UAVs to navigate safely through cluttered or fast-changing scenes. Additionally, these systems can support UAV detection and tracking, enhancing airspace management and security. Furthermore, event-based vision can facilitate object recognition in aerial imagery, such as identifying landmarks, vehicles, or other UAVs, even under challenging conditions like low light or high-speed motion. By leveraging the findings of this study, such applications could benefit from improved robustness and efficiency, ensuring reliable operation in the demanding environments often encountered by UAVs.\\

The study is limited by the availability of large-scale datasets for pre-training ViTs, which typically require extensive data to achieve optimal performance. Moreover, the literature on event representations and noise modeling for event-based data is still emerging, posing additional challenges. Despite these limitations, this research contributes to the growing field of event-based vision by highlighting the capabilities of ViT models and their potential application in scenarios where noise resilience is critical. The findings of this study not only provide insights into the strengths and weaknesses of these models but also lay the groundwork for their future implementation in real-world environments, such as UAV-based systems where robust performance under noisy conditions is essential.

\section{Methodology}

This section outlines the dataset used, the process for video reconstruction and event representation, the architectures of the models evaluated, and the evaluation procedure under various noise conditions. The evaluation framework for performance comparison is also detailed.

\subsection{Dataset}

\begin{figure}[htp!]
    \centering
    \begin{subfigure}{0.3\textwidth}
        \includegraphics[width=\linewidth]{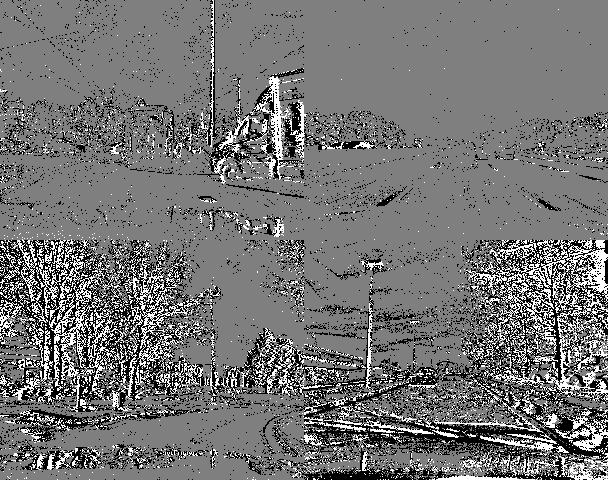}
        \caption{Training dataset sample.}
        \label{fig:gen1_train_ds}
    \end{subfigure}
    \hfill
    \begin{subfigure}{0.3\textwidth}
        \includegraphics[width=\linewidth]{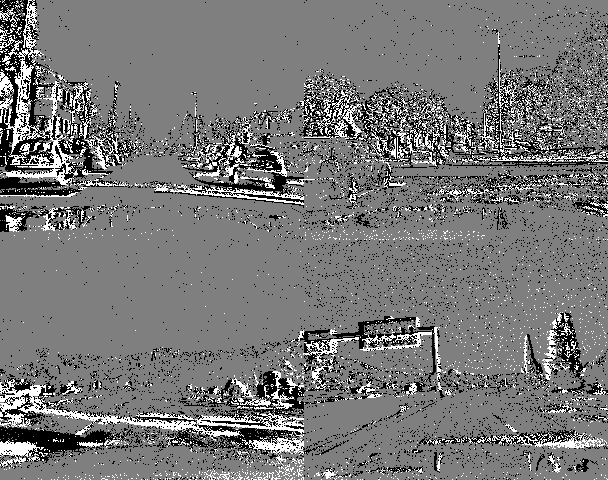}
        \caption{Validation dataset sample.}
        \label{fig:gen1_val_ds}
    \end{subfigure}
    \hfill
    \begin{subfigure}{0.3\textwidth}
        \includegraphics[width=\linewidth]{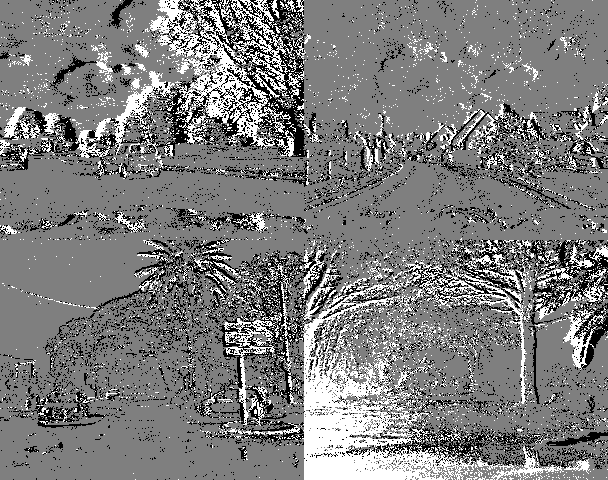}
        \caption{Testing dataset sample.}
        \label{fig:gen1_test_ds}
    \end{subfigure}
    
    \caption{GEN1 dataset}
    \label{fig:gen1_ds}
\end{figure}

The GEN1 dataset, developed specifically for automotive applications, is utilized in this study. In Figure \ref{fig:gen1_ds} can be seen some examples of its content. It consists of asynchronous events captured using an Asynchronous Time-based Image Sensor (ATIS) camera, which records changes in brightness rather than full image frames. The dataset includes two primary classes, cars and pedestrians, recorded under varying conditions, providing a diverse set of event streams for training and evaluation. These event streams include spatial coordinates, timestamps, and polarity information, making them suitable for both frame-based and event-based deep learning representations.

\subsection{Video Reconstruction}

In event-based vision systems, raw event streams are sparse and unstructured, making them unsuitable for direct input into most deep learning architectures. To prepare the raw event data for deep learning, a video reconstruction step was implemented. Events were accumulated over predefined temporal windows to create event frames, where each frame captured a snapshot of activity within its respective time window. Motion correction techniques, such as optical flow estimation and sensor calibration, were employed to mitigate distortions caused by high-speed movements or sensor shifts, ensuring that the reconstructed frames accurately reflected the underlying dynamics of the scene. The reconstruction transforms the data from raw event streams to structured event frames. Additionally, the frames were extracted and organized according to their respective classes, a crucial preprocessing step for effective model implementation.



\subsection{Event Representation Pipeline}

\begin{figure}[htp!]
    \centering
    \rotatebox{90}{
        \includegraphics[width=0.3\linewidth]{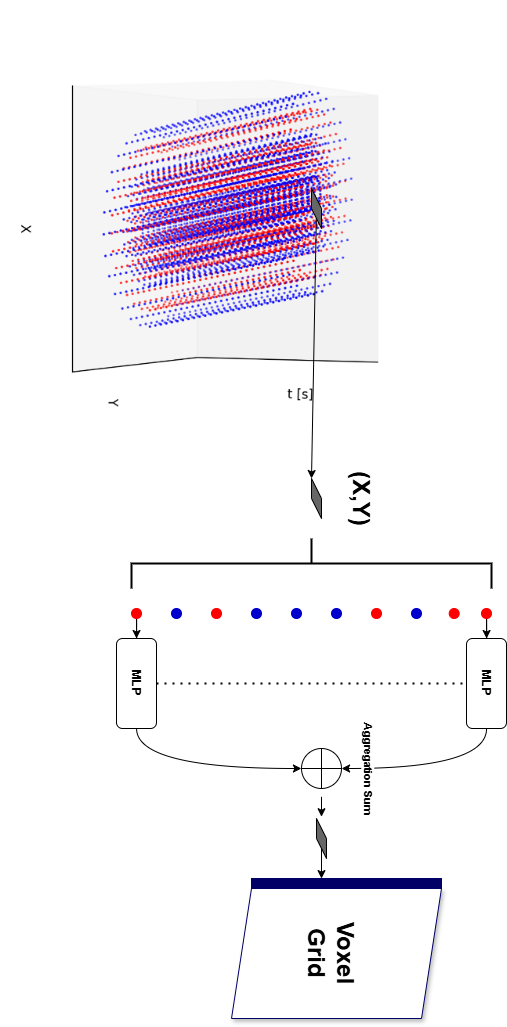}
    }
    \caption{EST representation pipeline \cite{gehrig2019end,cannici2020differentiable}.}
    \label{fig:EST representation pipeline}
\end{figure}

The raw asynchronous events are transformed into a structured format using the Event Spike Tensor (EST) representation \cite{gehrig2019end} shown in Figure \ref{fig:EST representation pipeline}. This involves mapping events into a multidimensional tensor that encodes spatial, temporal, and polarity information. The tensor serves as an input for deep learning models by aggregating events into a grid-based structure that retains both temporal resolution and spatial locality. This representation allows for effective feature extraction and model training by preserving fine-grained temporal details while maintaining spatial coherence, which is crucial for detecting fast-moving or subtle changes in the scene. Additionally, the EST structure facilitates efficient learning by reducing the complexity of raw event data, making it more suitable for deep learning models to capture relevant patterns and dynamics.

\subsection{Deep Learning Architectures}

\begin{figure}[htp!]
    \centering
    \includegraphics[width=0.4\linewidth]{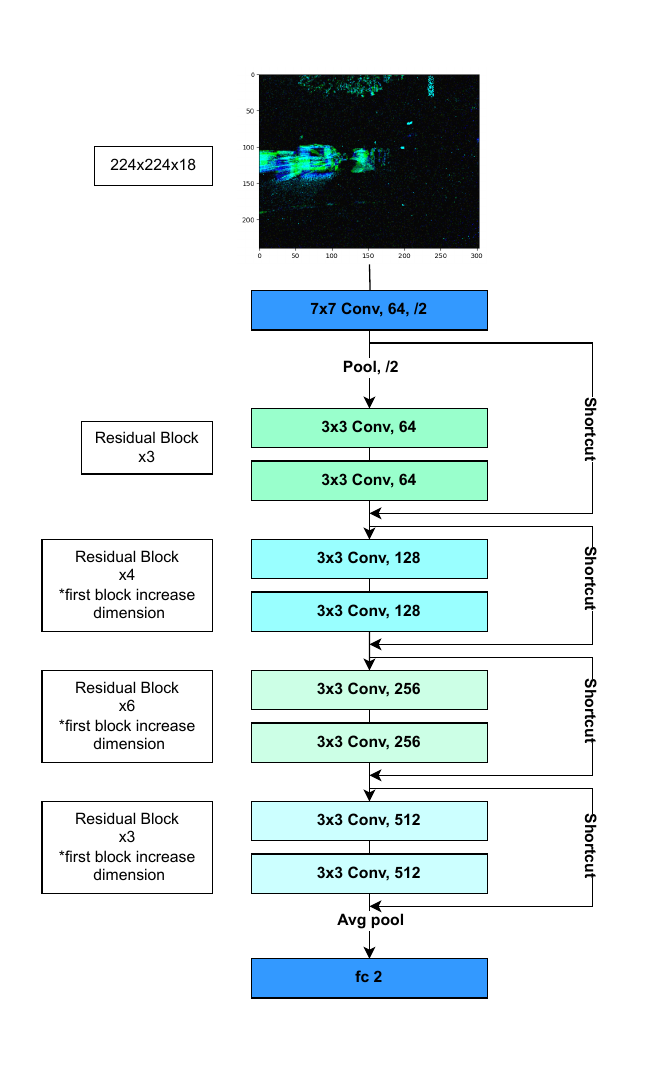}
    \caption{ResNet34 Architecture \cite{he2016deep,zhang2023multi}.}
    \label{fig:ResNet34 Architecture}
\end{figure}

\begin{figure}[htp!]
    \centering
    \includegraphics[width=0.6\linewidth, angle=90]{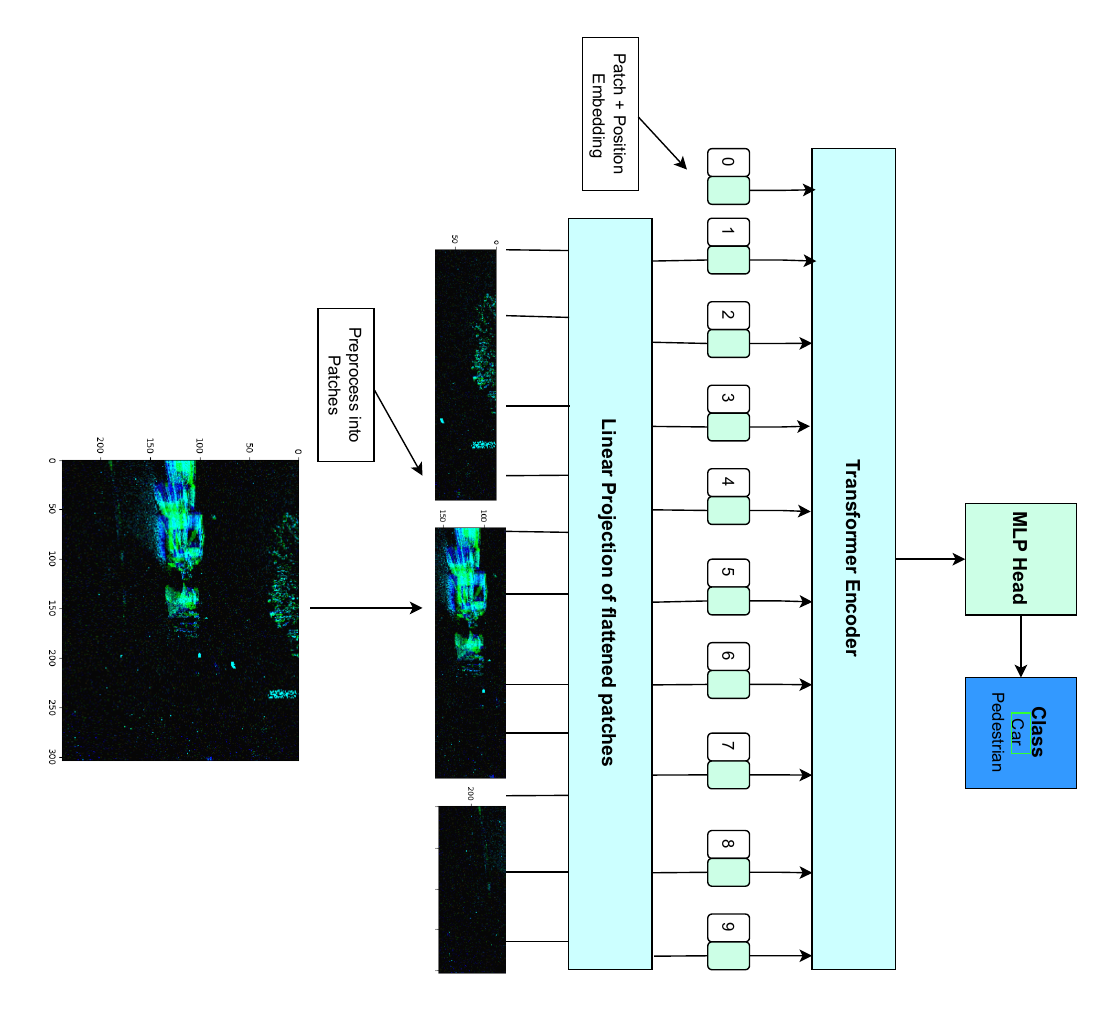}
    \caption{Vision Transformer main blocks architecture.}
    \label{fig:Vision Transformer main blocks architecture}
\end{figure}

Two deep learning models, ResNet34 and Vision Transformer B16, are selected for evaluation. ResNet34, whose architecture is shown in Figure \ref{fig:ResNet34 Architecture}, is a well-established CNN that employs residual connections to facilitate the training of deep architectures. ViT B16, whose architecture is shown in Figure \ref{fig:Vision Transformer main blocks architecture} is a transformer-based model that divides input frames into fixed-size patches and processes them as sequences using a multi-head self-attention mechanism. Both models are fine-tuned in this work on the GEN1 dataset, optimizing their hyperparameters to enhance their performance on event-based data.

\subsection{Evaluation on Added Noise}

Noise was introduced into the GEN1 dataset to simulate real-world challenges often encountered in event-based vision systems. The types of noise studied include spatial shifts of events, loss of events, and polarity reversals. Each type of noise was applied at varying intensities to assess its impact on model performance.

\subsubsection{Events Shift}

Spatial shifting of events simulates errors in sensor calibration or motion artifacts, where the recorded event positions are displaced along the \(x\)- and \(y\)-axes. For each event \(e_i = (x_i, y_i, t_i, p_i)\), where $x_i$, $y_i$, and $t_i$ are the spatiotemporal coordinates and $p_i$ the polarity, the spatial coordinates \((x_i, y_i)\) are shifted by offsets \(\Delta x\) and \(\Delta y\):

\begin{equation}
x_i' = x_i + \Delta x, \quad y_i' = y_i + \Delta y
\end{equation}

Here, \((x_i', y_i')\) are the new coordinates after the shift, and \(\Delta x\) and \(\Delta y\) are sampled from a uniform distribution:

\begin{equation}
\Delta x, \Delta y \sim \mathcal{U}(-\delta, \delta)
\end{equation}

where \(\delta\) is the maximum shift magnitude. The shifted events are evaluated for various values of \(\delta\), allowing analysis of the models’ robustness to spatial distortions.

\subsubsection{Events Loss}\label{sec:event_loss}

Event loss simulates incomplete data caused by hardware limitations or environmental conditions. A subset of events is randomly removed from the event stream. For an original event stream containing \(N\) events, the modified stream retains only a fraction \(1 - \eta\), where \(\eta\) represents the event loss rate.

The resulting event set \(E'\) is defined as:

\begin{equation}
E' = \{e_i \in E \mid r_i > \eta \}, \quad r_i \sim \mathcal{U}(0, 1)
\end{equation}

Here, \(r_i\) is a random number uniformly sampled between 0 and 1 for each event \(e_i\). Events are removed if \(r_i \leq \eta\). Different values of \(\eta\) are tested to evaluate the models under increasing data loss.

\subsubsection{Polarity Reversal}

Polarity reversal introduces errors in the polarity of events, which indicate the direction of brightness changes. For each event \(e_i = (x_i, y_i, t_i, p_i)\), the polarity \(p_i \in \{-1, +1\}\) is flipped with a probability \(\rho\):

\begin{equation}
p_i' =
\begin{cases}
-p_i, & \text{if } r_i \leq \rho \\
p_i, & \text{otherwise}
\end{cases}
\end{equation}

Here, \(p_i'\) is the new polarity, and \(r_i \sim \mathcal{U}(0, 1)\) is a random number sampled for each event. The parameter \(\rho\) controls the fraction of events with reversed polarity, simulating noise in event encoding.\\

Each noise type was applied independently to the dataset going from 5\% to 20\%, enabling a comprehensive evaluation of model robustness. These formulations ensure that the noise is introduced systematically, maintaining control over the intensity and allowing meaningful comparisons across different noise scenarios.

\subsection{Training}

The training procedure for ResNet34 and ViT B16 was designed to ensure a fair comparison while optimizing the models for event-based object classification. The details of the training process are as follows:

\begin{itemize} 
    \item \textbf{Preprocessing:}
    The raw event data was preprocessed into event frames using the EST representation. This method preserved temporal resolution and spatial locality.\\

    \item \textbf{Data Splitting:}  
    The GEN1 dataset was split into training, validation, and testing subsets in a ratio of 70:15:15. This ensured sufficient data for training while maintaining a balanced validation set for hyperparameter tuning and a test set for performance evaluation.\\
    
    \item \textbf{Model Initialization and Fine-Tuning:}  
    Both ResNet34\footnote{\url{https://pytorch.org/vision/main/models/generated/torchvision.models.resnet34.html}} and ViT B16\footnote{\url{https://pytorch.org/vision/main/models/generated/torchvision.models.vit_b_16.html}} were initialized with weights pre-trained on ImageNet \cite{russakovsky2015imagenet}. The first and last layers of ResNet34 were adapted to accommodate the EST input and the binary classification task. For ViT B16, the positional encodings and class token were fine-tuned to suit the event-based dataset.\\
    
    \item \textbf{Optimization and Hyperparameters:}  
    The Adam optimizer with a learning rate of \(1 \times 10^{-4}\) was employed for both models. Training was conducted for 50 epochs, with early stopping based on validation loss to prevent overfitting. A batch size of 8 was selected to balance memory efficiency and computational speed.\\
    
    \item \textbf{Noise Evaluation:}  
    Noise was systematically added to the training and validation sets to simulate real-world scenarios. Noise types included coordinates shift, event loss, and polarity reversal, applied at varying levels, as described in Section \ref{sec:event_loss}. This step evaluated the robustness of each model under challenging conditions.\\
    
    \item \textbf{Evaluation During Training:}  
    Performance metrics, including accuracy, loss, and Area Under the Curve (AUC), were monitored on the validation set at the end of each epoch. This ensured consistent tracking of model improvements and highlighted any overfitting trends.\\
    
    \item \textbf{Hardware and Frameworks:}  
    Training experiments were conducted on an NVIDIA GPU-based high-performance cluster using PyTorch. Supporting libraries were utilized for event-based vision processing and representation.
\end{itemize}

\subsection{Performance Metrics}

To evaluate and compare the performance of ResNet34 and ViT B16, several performance metrics were employed. These metrics provide a comprehensive understanding of the models' accuracy, robustness, and reliability under varying noise conditions. The metrics used are as follows:\\

\begin{itemize}
    \item \textbf{Accuracy}: Measures the proportion of correctly predicted instances out of the total cases. It is calculated using Equation \ref{eq:accuracy}.
    
    \begin{equation}
    \text{Accuracy} = \frac{TP + TN}{TP + TN + FP + FN}
    \label{eq:accuracy}
    \end{equation}

     \noindent where \(TP\) represents true positives, \(TN\) true negatives, \(FP\) false positives, and \(FN\) false negatives. However, accuracy alone can be misleading, especially in cases with imbalanced data.\\
    
    \item \textbf{Receiver Operating Characteristic (ROC) curve and Area Under the Curve (AUC)}: The ROC curve plots the True Positive Rate (TPR) against the False Positive Rate (FPR) at various threshold settings for classifying the data as one class or the other. The AUC is a single scalar value that summarizes the overall performance of the model, with higher values indicating better discriminatory ability \cite{herlocker2004evaluating}.\\

    \item \textbf{Precision-Recall (PR) curve and Average Precision (AP)}: The PR curve evaluates the trade-off between precision (positive predictive value) and recall (sensitivity) across different thresholds. Average Precision (AP) quantifies the area under the PR curve, highlighting the model's performance in imbalanced datasets where precision and recall are critical \cite{saito2015precision,sofaer2019area}.

    \begin{equation}
    \text{Precision} = \frac{TP}{TP + FP}
    \label{eq:precision}
    \end{equation}
    
    \begin{equation}
    \text{Recall} = \frac{TP}{TP + FN}
    \label{eq:recall}
    \end{equation}\\

    \item \textbf{F1 score} is the harmonic mean of precision and recall, offering a more balanced measure compared to accuracy in such scenarios \cite{TALHA2023126881}. It is calculated as shown in Equation \ref{eq:recall_P}.

     \begin{equation}
    F1 = 2 \times \frac{\text{Precision} \times \text{Recall}}{\text{Precision} + \text{Recall}}
    \label{eq:recall_P}
    \end{equation}

\end{itemize}

\subsection{Evaluation Framework}

\begin{figure}[htp!]
    \centering
    \includegraphics[width=0.5\linewidth]{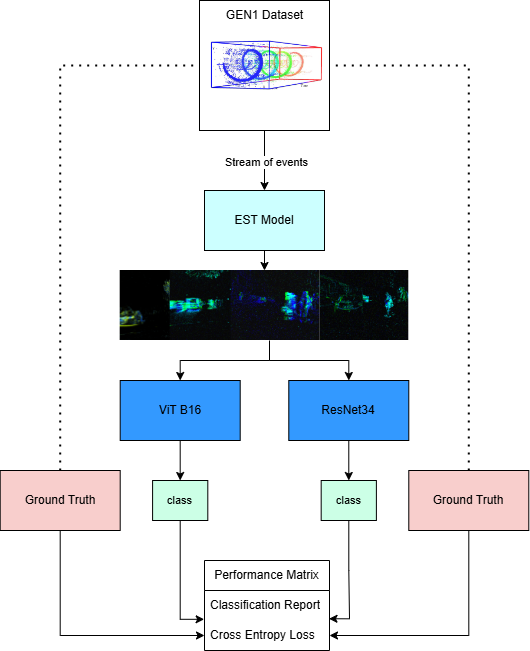}
    \caption{Classification evaluation procedure.}
    \label{fig:Classification evaluation procedure}
\end{figure}

The models are evaluated on the GEN1 dataset under both clean and noisy conditions using metrics such as accuracy, Receiver Operating Characteristic (ROC) curves, and Precision-Recall (PR) curves. For each noise type, the models' performance is recorded across multiple noise levels to analyze their robustness. Comparative analysis highlights the relative strengths and weaknesses of ResNet34 and ViT B16 in handling noise, providing insights into their suitability for real-world applications. Figure \ref{fig:Classification evaluation procedure} summarizes the evaluation framework used in this work.

\section{Results and Discussion}

\subsection{ResNet34: Clean Dataset}

ResNet34, when evaluated on the clean GEN1 dataset, achieved an accuracy of 88\%. These results demonstrate that ResNet34 effectively leverages its convolutional architecture to extract spatial features from event-based data. The model's parameters and training parameters are shown in Tables \ref{table:resnet_parameters_gen1} and \ref{table:resnet_training_parameters_gen1}, respectively.

\begin{table}[H]
\centering
\caption{ResNet34 model parameters for fine-tuning on the GEN1 dataset.}
\begin{tabular}{@{}>{\bfseries}l m{5cm}@{}}
\toprule
Parameter & Value \\ \midrule
Voxel Dimension & (9, 240, 304) \\ 
Crop Dimension & (18, 224, 224) \\ 
Number of Classes & 2 \\ 
MLP Layers & [1, 30, 30, 1] \\ 
Activation & LeakyReLU (negative slope=0.1) \\ 
Pretrained & True \\ \bottomrule
\end{tabular}
\label{table:resnet_parameters_gen1}
\end{table}

\begin{table}[H]
\centering
\caption{ResNet34 training parameters for fine-tuning on the GEN1 dataset.}
\begin{tabular}{@{}>{\bfseries}l m{5cm}@{}}
\toprule
Parameter & Value \\ \midrule
Batch Size & 8 \\ 
Number of folds & 5\\
Total Number of Epochs & 35 \\ 
Learning Rate & 2.2e-5 \\ 
Optimizer & Adam \\ 
Learning Rate Scheduler & ExponentialLR (gamma=0.34)
\\ \bottomrule
\end{tabular}
\label{table:resnet_training_parameters_gen1}
\end{table}

\begin{figure}[htp!]
    \centering
    \begin{minipage}{0.48\textwidth}
        \centering
        \includegraphics[width=\linewidth]{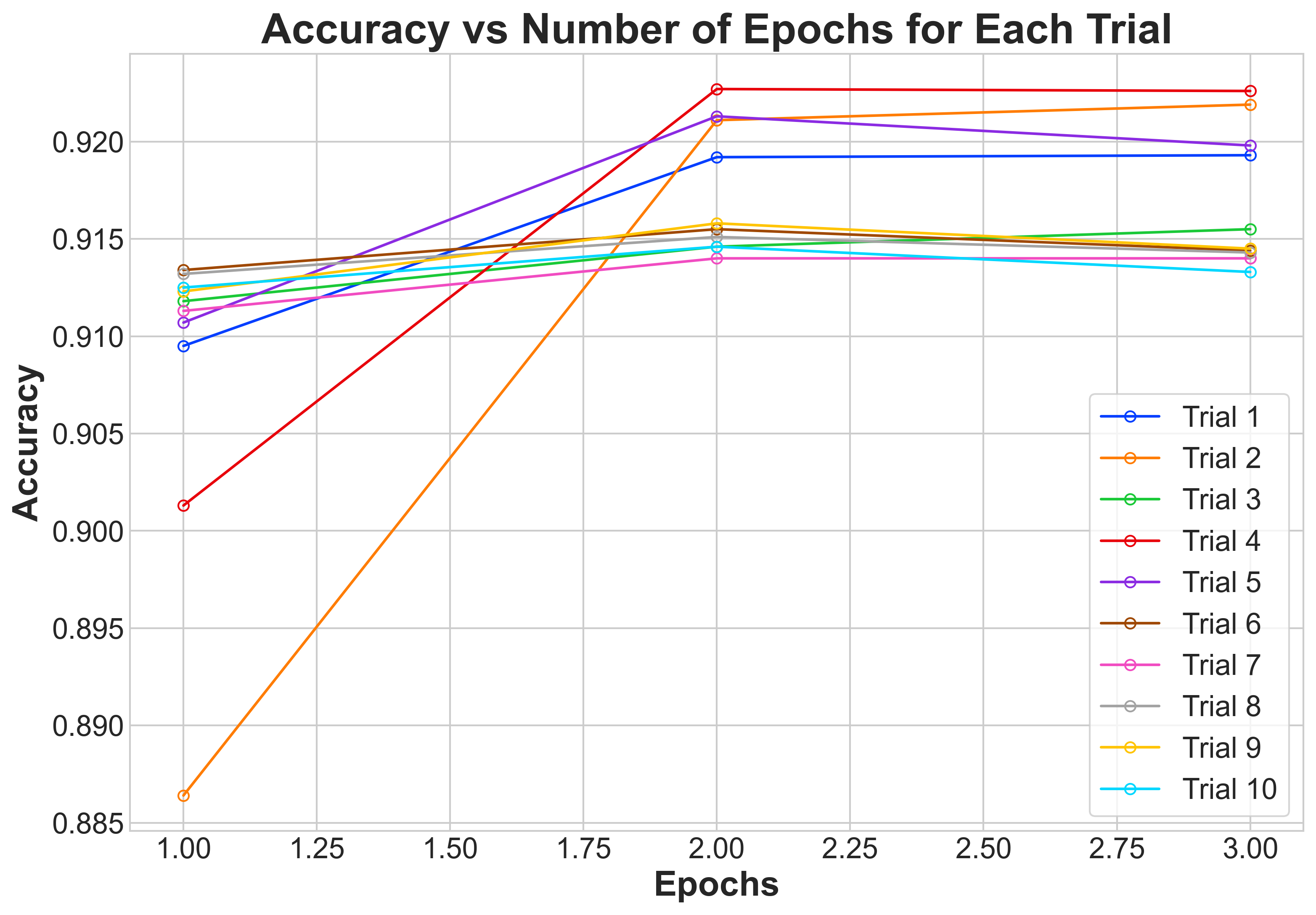}
        \caption{Study of the change of accuracy using different learning rates.}
        \label{fig:resnet_optuna_lr_acc}
    \end{minipage}\hfill
    \begin{minipage}{0.48\textwidth}
        \centering
        \includegraphics[width=\linewidth]{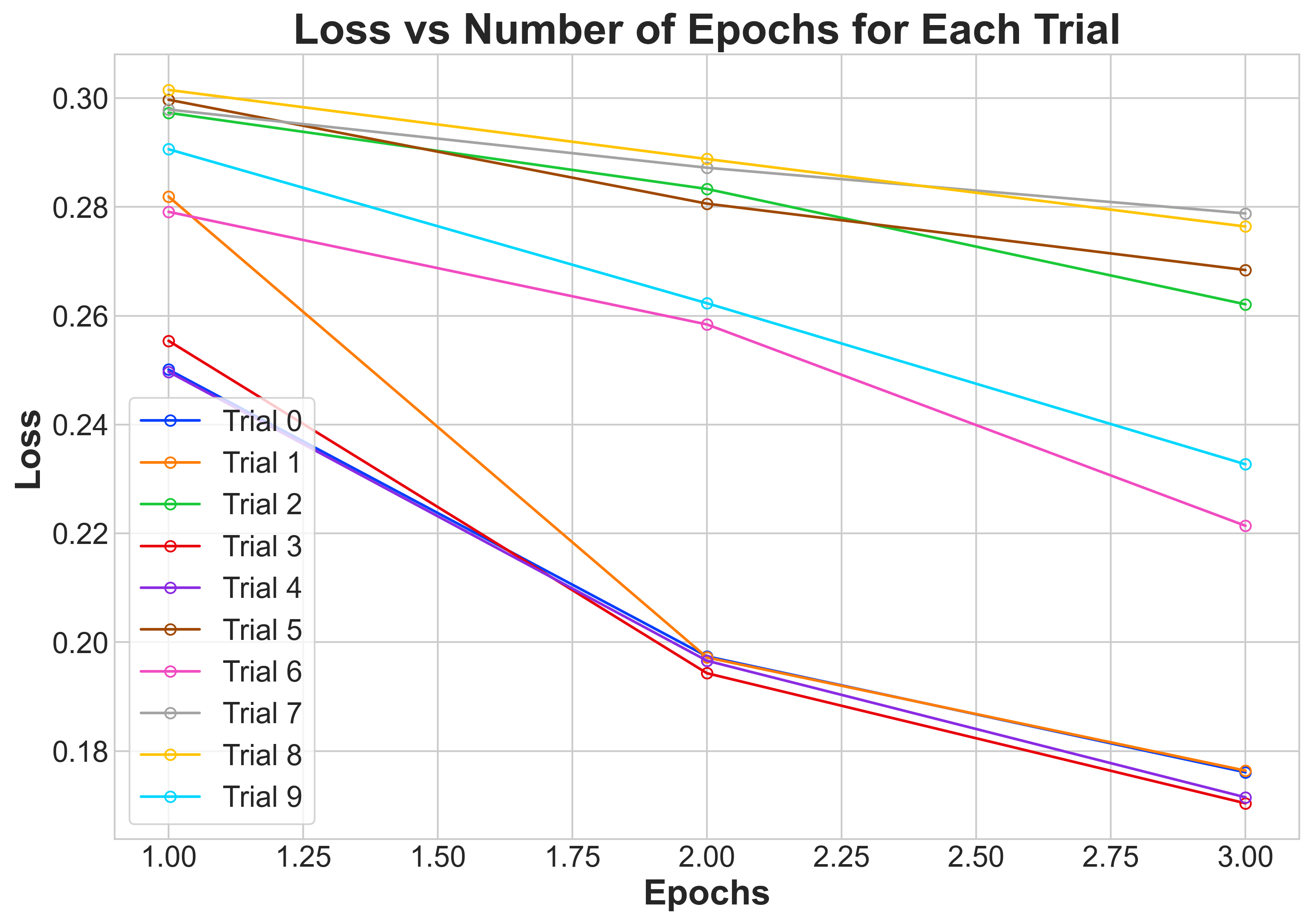}
        \caption{Study of the change of loss using different learning rates.}
        \label{fig:resnet_optuna_lr_loss}
    \end{minipage}
\end{figure}

The learning rate was determined after conducting a series of exploratory trials using Optuna\footnote{\url{https://optuna.org/}} to monitor performance variations in response to changes in the learning rate. Figures \ref{fig:resnet_optuna_lr_acc} and \ref{fig:resnet_optuna_lr_loss} display the outcomes from these Optuna trials, where the model underwent 10 trials, each consisting of 3 epochs. Among these, Trial 4 yielded the most promising results, identifying a learning rate of 2.2e-5 as optimal.

\begin{figure}[htp!]
    \centering
    \includegraphics[width=0.8\linewidth]{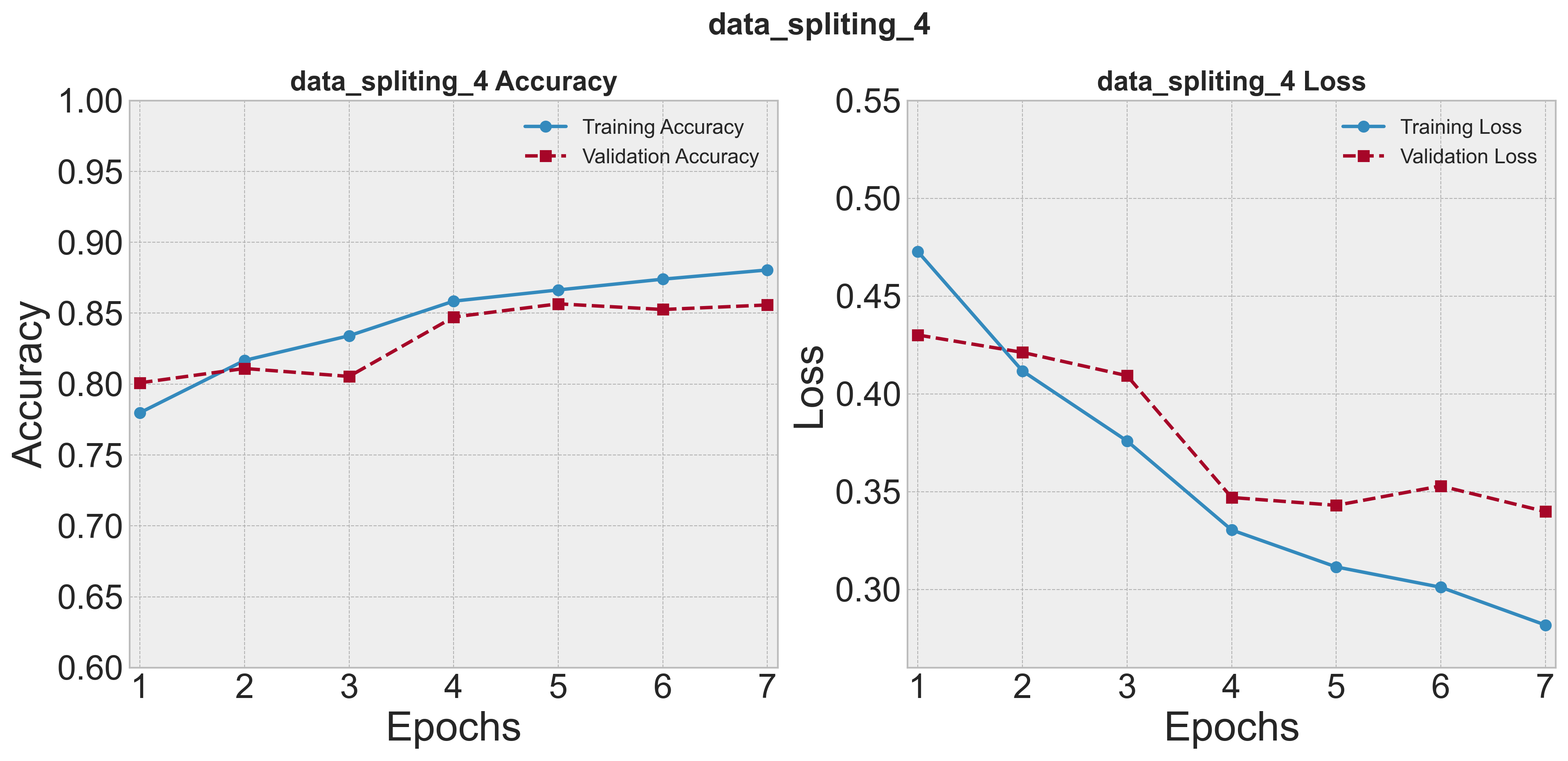}
    \caption[ResNet34 cross-validation: data\_splitting\_4.]{Training accuracy (left), and training loss (right)}
    \label{fig:fold4_train_vaal_curve}
\end{figure}

Cross-validation techniques were applied to ensure the best results while reducing any possible bias from the data splitting. The data was split into five folds. The best training and validation curves, corresponding to fold 4, are shown in Figure \ref{fig:fold4_train_vaal_curve}, the remaining folds are of similar behavior. The training accuracy, in the left plot, shows a gradual increase across the epochs, while the validation accuracy remains relatively stable after an initial rise, suggesting that the model’s performance on unseen data does not improve significantly after a certain point. The training loss, in the right plot, decreases consistently, but the validation loss exhibits fluctuations, especially in the later epochs, indicating a possible overfitting trend where the model performs well on training data but struggles on validation data.

\begin{figure}[htp!]
    \centering
    \includegraphics[width=0.9\linewidth]{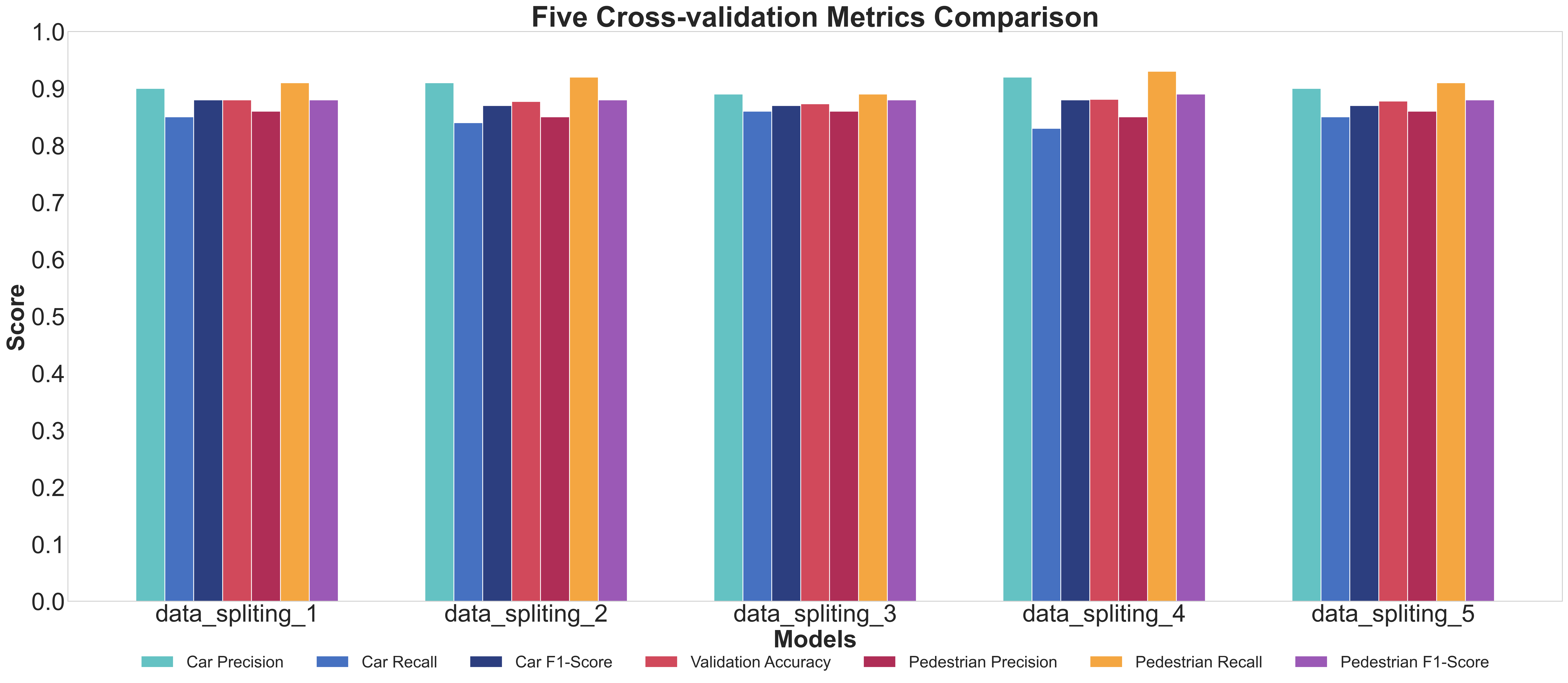}
    \caption{Comparison between the five folds of the ResNet34 models.}
    \label{fig:Comparison between the five folds}
\end{figure}

The comparative performance of these models trained with different folds is illustrated in Figure \ref{fig:Comparison between the five folds}. Although all models showed similar effectiveness, model 4 was selected for further testing on the noise dataset due to its superior performance.

\begin{figure}[htp!]
        \centering
        \includegraphics[width=0.5\linewidth]{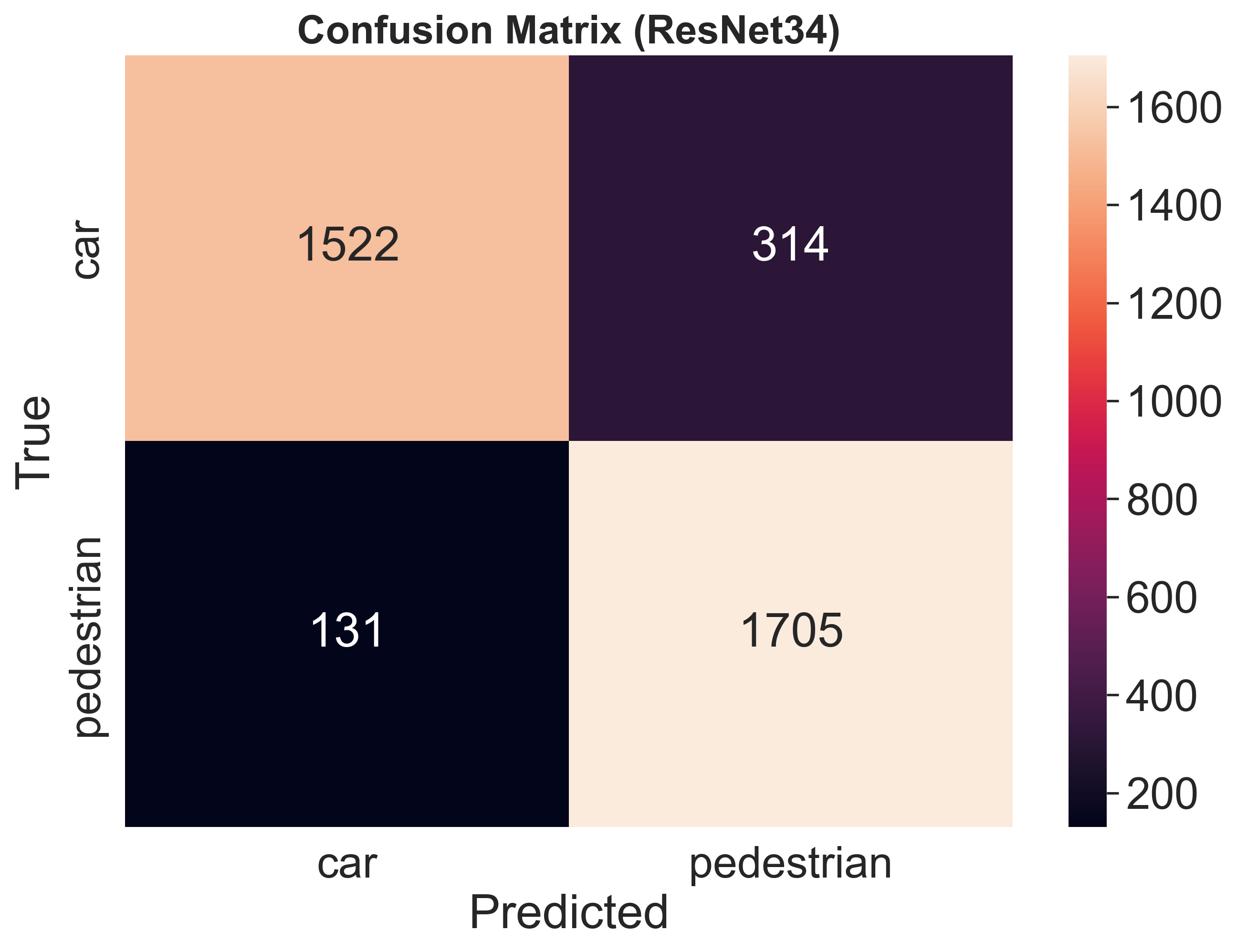}
        \caption{ResNet34 confusion matrix on the GEN1 dataset.}
        \label{fig:confusion_matrix_resnet}
    \end{figure}

\begin{figure}[htp!]
        \centering
        \includegraphics[width=0.5\linewidth]{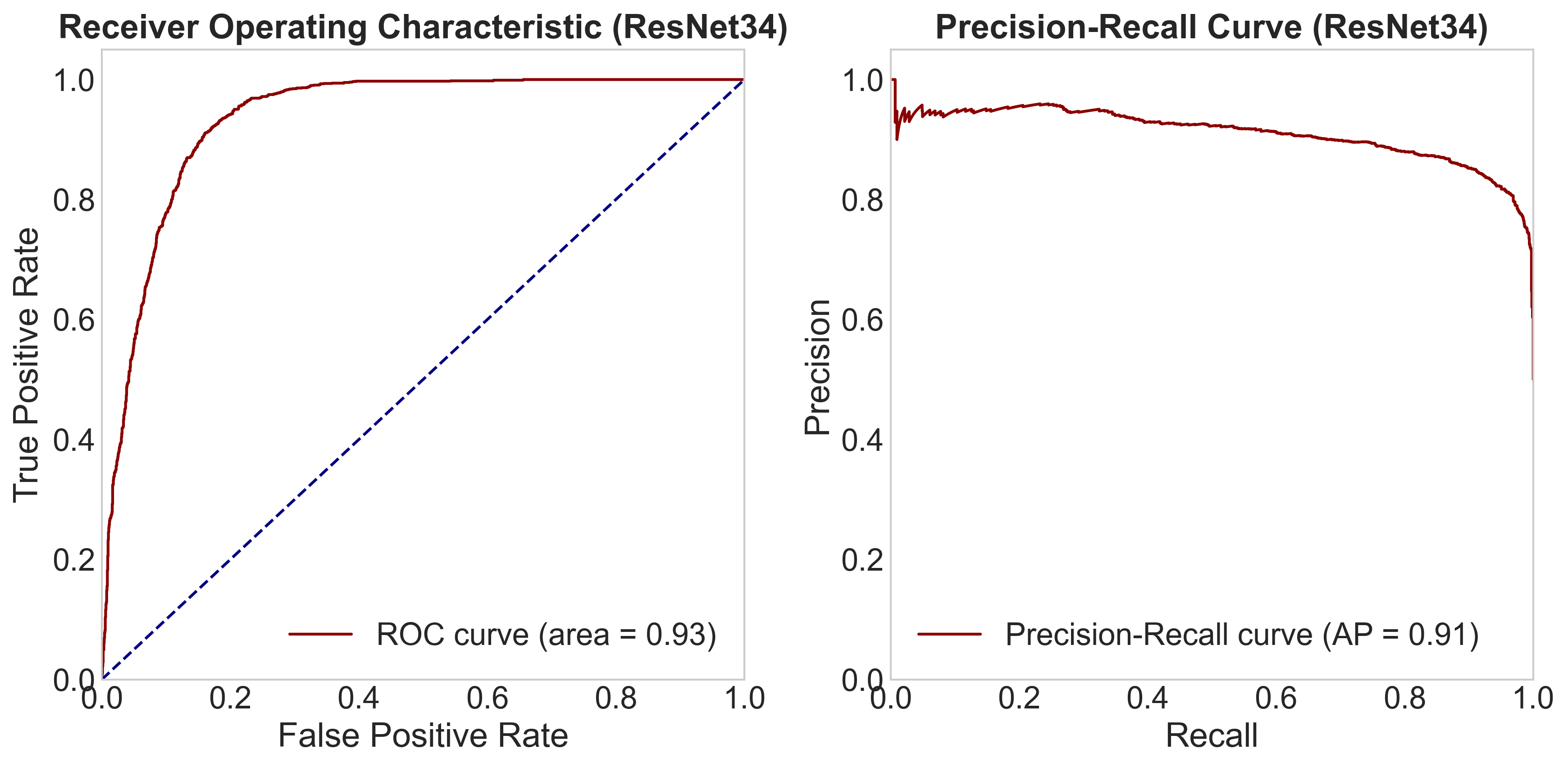}
        \caption{ResNet34 ROC and PR curves on the GEN1 dataset.}
        \label{fig:ROC_PR_curve_resnet34}
    \end{figure}

The confusion matrix, the ROC, and the PR curve for the ResNet34 are shown in Figures \ref{fig:confusion_matrix_resnet} and \ref{fig:ROC_PR_curve_resnet34}. The ResNet34 model correctly identified 82.9\% of car instances and 92.9\% of pedestrian instances, while misclassifying 17.1\% of cars as pedestrians and 7.1\% of pedestrians as cars. The ROC curve reveals an Area Under the Curve (AUC) of 0.93 and an Average Precision (AP) of 0.91.

The model's performance was evaluated, with the resulting accuracy and loss detailed in Table \ref{table:resnte_validation_metrics_gen1}. These outcomes serve as benchmarks for comparing the model's performance with that of the ViT model.\\

\begin{table}[H]
\centering
\caption{ResNet34 classification report on the GEN1 dataset.}
\label{table:resnte_validation_metrics_gen1}
\begin{tabular}{lcccc}
\hline
\textbf{Class}      & \textbf{Precision} & \textbf{Recall} & \textbf{F1-Score} & \textbf{Support} \\
\hline
Car          & 0.92 & 0.83 & 0.87 & 1836 \\
Pedestrian   & 0.84 & 0.93 & 0.88 & 1836 \\
\hline
\textbf{Accuracy}      &       &       & 0.88 & 3672 \\
\textbf{Macro Avg}     & 0.88 & 0.88 & 0.88 & 3672 \\
\textbf{Weighted Avg}  & 0.88 & 0.88 & 0.88 & 3672 \\
\hline
\end{tabular}
\end{table}

\subsection{ViT B16: Clean Dataset}

On the clean GEN1 dataset, ViT B16 achieved an accuracy of 86\%. Despite a marginally lower classification accuracy compared to ResNet34, the model exhibited competitive results. The ViT B16's transformer architecture, which processes input as sequential patches, showed robust performance, particularly given its pre-training on smaller datasets. This highlights the potential of transformer-based models to generalize well on event-based data, even with limited training resources. The model's parameters and training parameters are shown in Tables \ref{table:vit_parameters_gen1} and \ref{table:vit_training_parameters_gen1}, respectively. 

\begin{table}[H]
\centering
\caption{ViT B16 model parameters for fine-tuning on the GEN1 dataset.}
\begin{tabular}{@{}>{\bfseries}l m{5cm}@{}}
\toprule
Parameter & Value \\ \midrule
Voxel Dimension & (9,240,304) \\ 
Crop Dimension & (18, 224, 224) \\
ViT Preprocessing& (3, 244, 244)\\
Number of Classes & 2 \\ 
MLP Layers & [1, 30, 30, 1] \\ 
Activation & LeakyReLU (negative slope=0.1) \\ 
Pretrained & True \\ \bottomrule
\end{tabular}
\label{table:vit_parameters_gen1}
\end{table}

\begin{table}[H]
\centering
\caption{ViT training parameters for fine-tuning on the GEN1 dataset.}
\begin{tabular}{@{}>{\bfseries}l m{5cm}@{}}
\hline
Parameter & Value \\ \hline
Batch Size & 10 \\ 
drop\_out layers & 0\\
Number of Epochs & 10 \\ 
Learning Rate & 2.2e-5 \\ 
Optimizer & Adam \\ 
Learning Rate Scheduler & ExponentialLR (gamma=0.34)
\\ \hline
\end{tabular}
\label{table:vit_training_parameters_gen1}
\end{table}

\begin{figure}[htp!]
    \centering
    \includegraphics[width=0.7\linewidth]{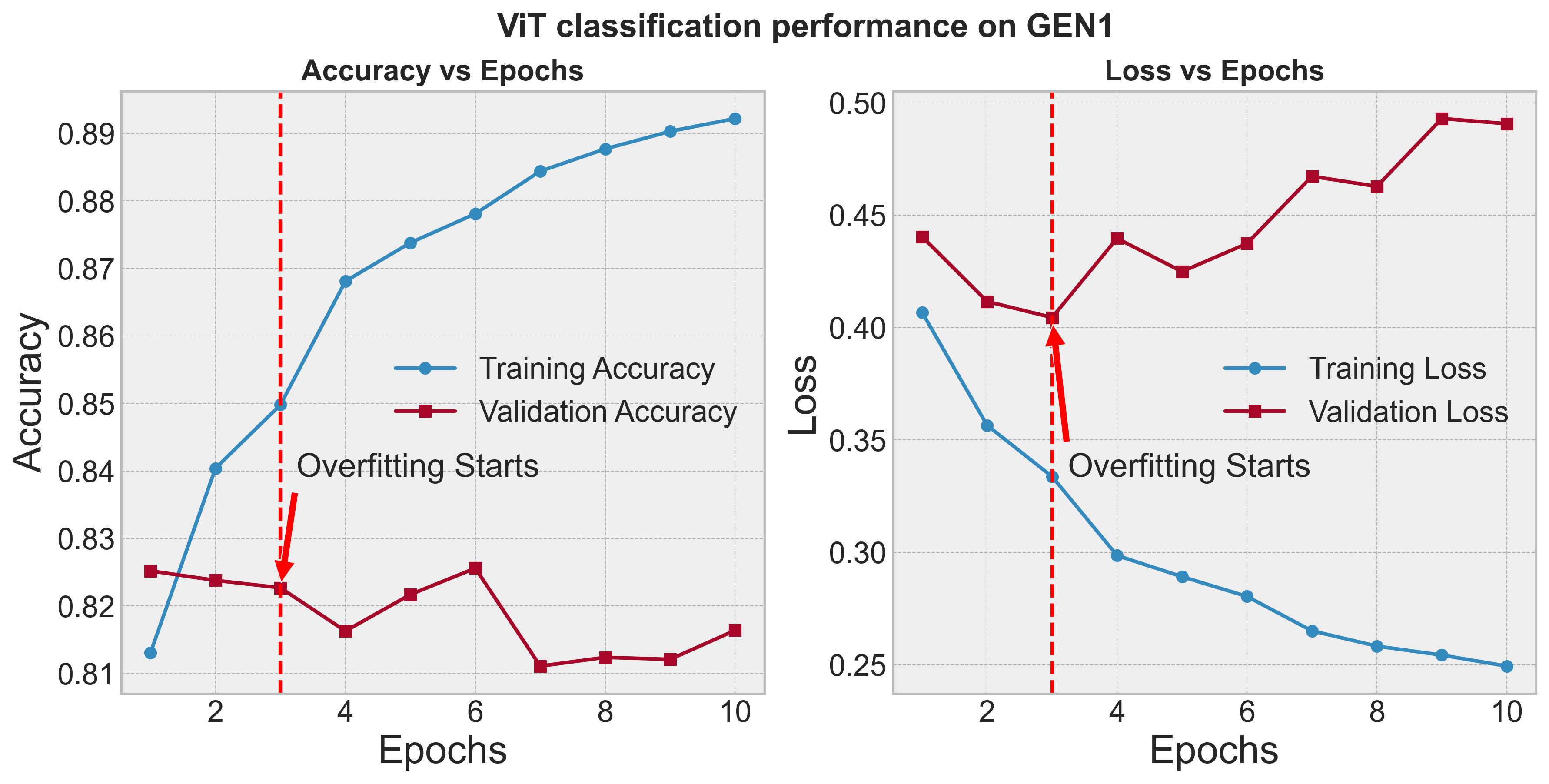}
    \caption{ViT B16 training and validation curves on GEN1 without dropout layer.}
    \label{fig:vit_gen1_performance}
\end{figure}

\begin{figure}[htp!]
    \centering
    \includegraphics[width=0.7\linewidth]{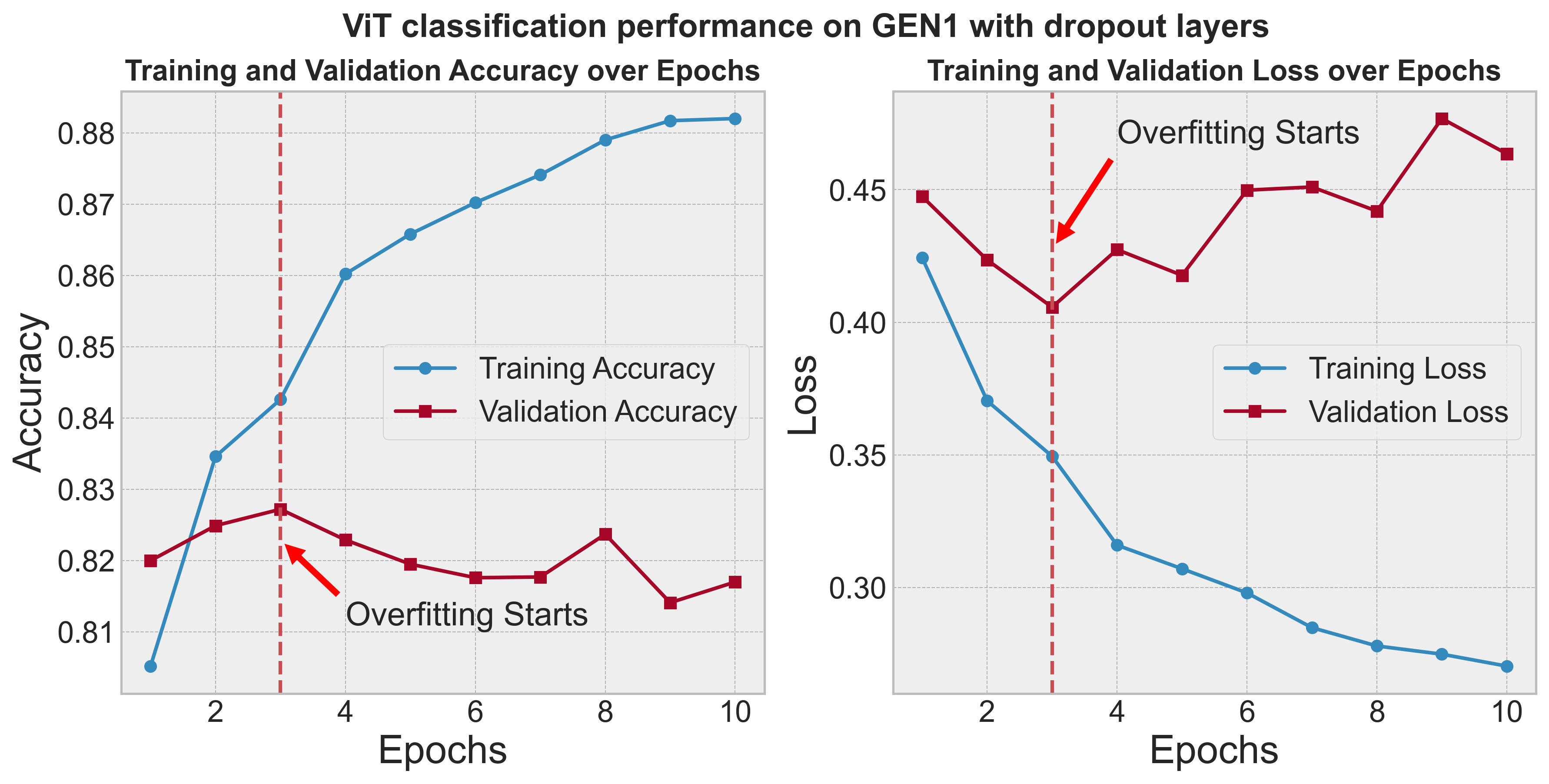}
    \caption{ViT B16 training and validation curves on GEN1 with dropout layer.}
    \label{fig:vit_gen1_performance_do}
\end{figure}

The model was tested with added dropout layer and without dropout layer to prevent overfitting. Figures \ref{fig:vit_gen1_performance} and \ref{fig:vit_gen1_performance_do} show the results of the validation sets. The accuracy and loss curves displayed in Figure \ref{fig:vit_gen1_performance_do} indicate that incorporating dropout layers slightly enhanced the model's performance. This outcome could be attributed to the dataset's size not being sufficiently large to fully benefit from the dropout layer, which often requires larger datasets to regularize the model effectively \cite{srivastava2014dropout}. 

\begin{figure}[htp!]
        \centering
        \includegraphics[width=0.5\linewidth]{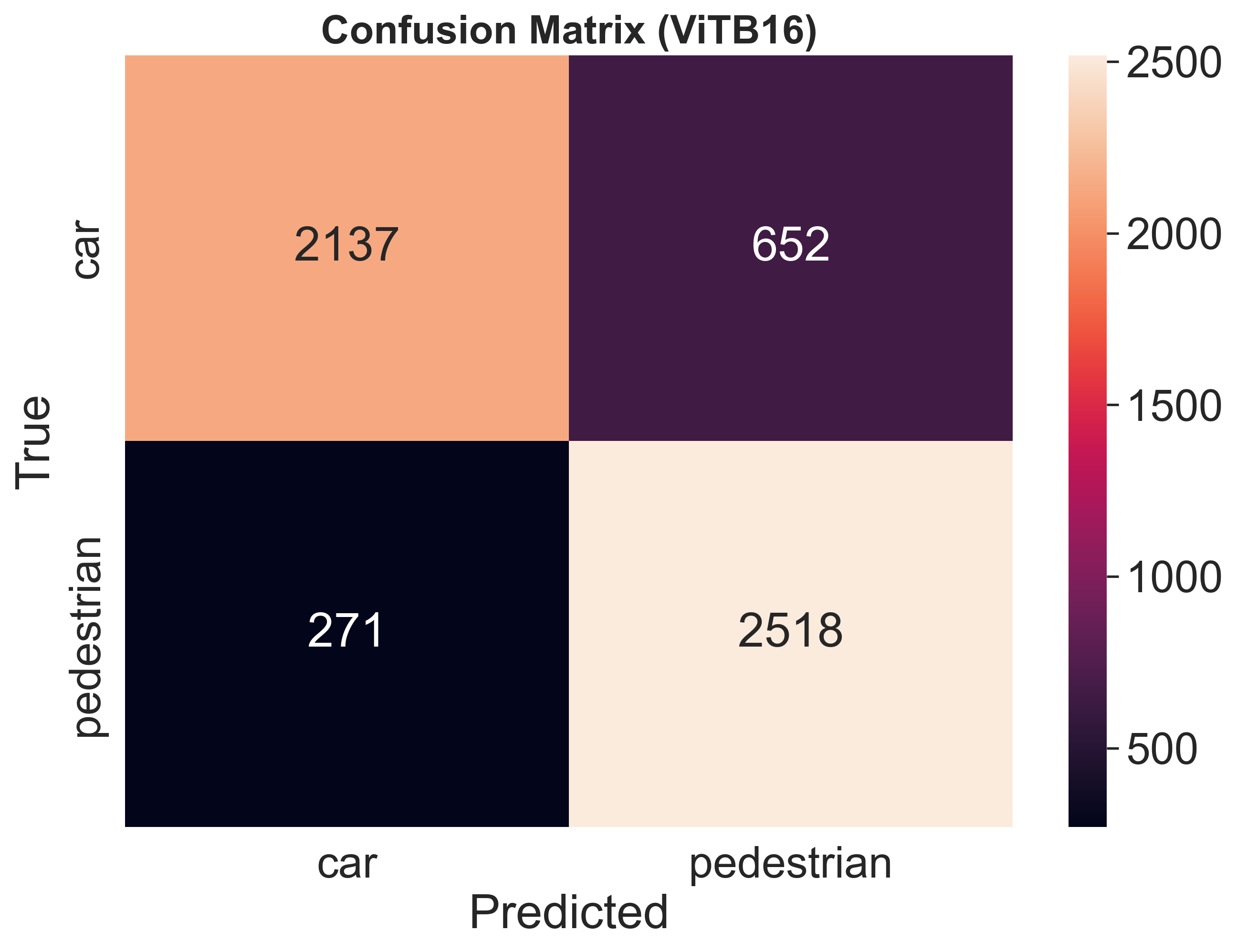}
        \caption{ViT B16 confusion matrix on the GEN1 dataset.}
        \label{fig:confusion_matrix_vit}
\end{figure}

\begin{figure}[htp!]
        \centering
        \includegraphics[width=0.5\linewidth]{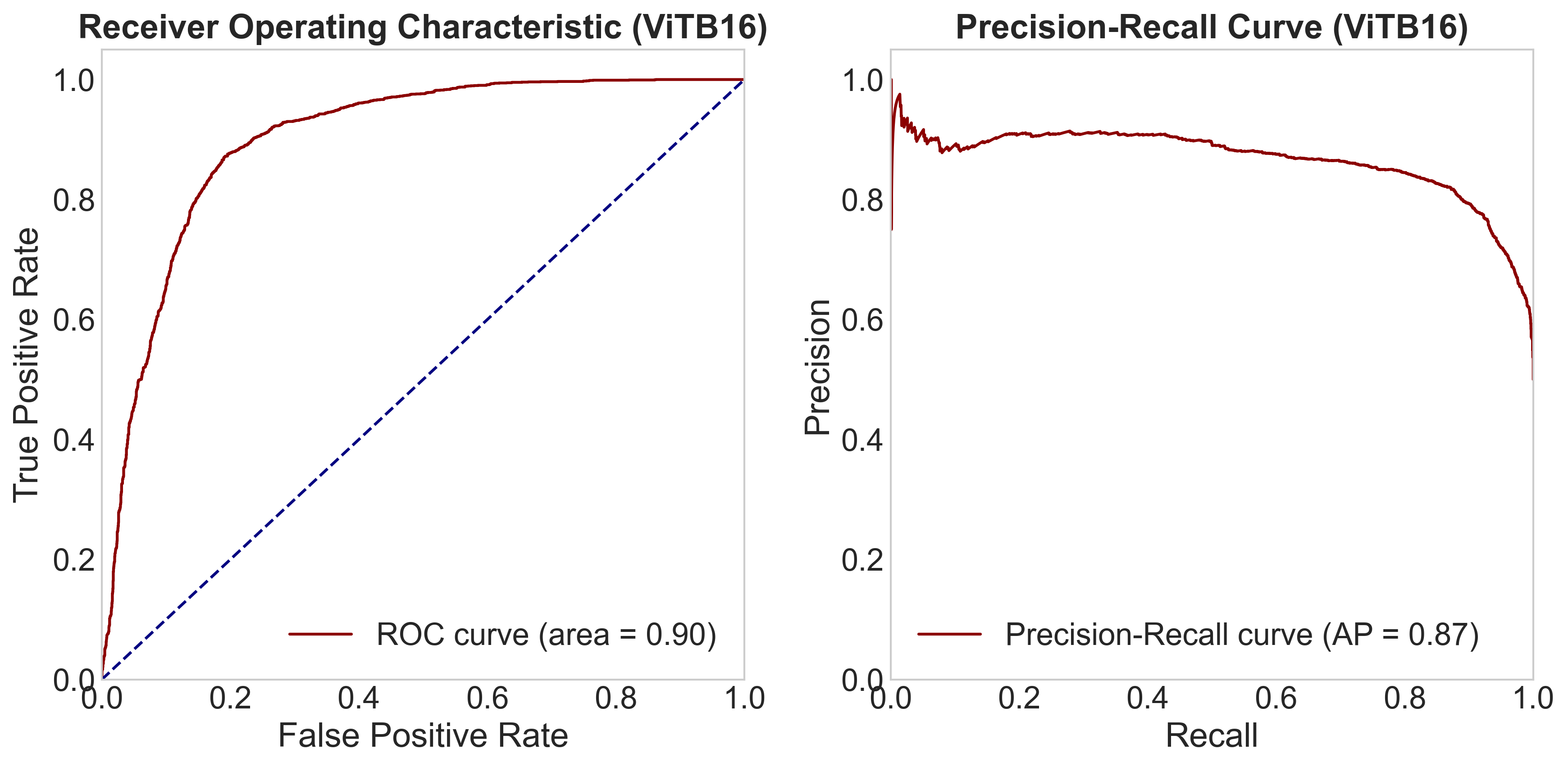}
        \caption{ViTB16 ROC and PR curves on the GEN1 dataset.}
        \label{fig:ROC_PR_curve_vit}
    \end{figure}

The confusion matrix, the ROC, and the PR curve for the ViT B16 are shown in Figures \ref{fig:confusion_matrix_vit} and \ref{fig:ROC_PR_curve_vit}. The model correctly classified 76.6\% of cars and 90.3\% of pedestrians, but it made more errors, misclassifying 23.4\% of cars as pedestrians and 9.7\% of pedestrians as cars. The ROC curve reveals an Area Under the Curve (AUC) of  0.90 and an Average Precision (AP) of 0.87.

The comparative test results for both models are presented in Table \ref{tab:vit_performance_2models}, with detailed classification results in Table \ref{tab:vit_do_layer_class_report}. The findings indicate that while both models exhibit similar performance levels, the model with dropout layers demonstrates marginally better results. Consequently, this model was chosen for further comparison against the best-performing ResNet34 model and additional testing with noise datasets.

\begin{table}[H]
\centering
\caption{Performance of ViT B16 with different dropout layer configurations.}
\label{tab:vit_performance_2models}
\begin{tabular}{@{}lcc@{}}
\hline
            & \textbf{Dropout Layers = 0} & \textbf{Dropout Layers = 2} \\ \hline
\textbf{Accuracy}  & 0.85                 & 0.86 \\
\textbf{Precision} & 0.86                 & 0.87 \\
\textbf{Recall}    & 0.85                 & 0.86\\
\textbf{F1-score}  & 0.85                 & 0.86 \\
\hline
\end{tabular}
\end{table}

\begin{table}[H]
\centering
\caption{VitB16 with dropout layers Classification Report for GEN1 dataset.}
\label{tab:vit_do_layer_class_report}
\begin{tabular}{lcccc}
\hline
\textbf{Class}      & \textbf{Precision} & \textbf{Recall} & \textbf{F1-Score} & \textbf{Support} \\
\hline
Car          & 0.92 & 0.78 & 0.85 & 3862 \\
Pedestrian   & 0.81 & 0.94 & 0.87 & 3858 \\
\hline
\textbf{Accuracy}      &       &       & 0.86 & 7720 \\
\textbf{Macro Avg}     & 0.87 & 0.86 & 0.86 & 7720 \\
\textbf{Weighted Avg}  & 0.87 & 0.86 & 0.86 & 7720 \\
\hline
\end{tabular}
\end{table}

\subsection{Evaluation on Noisy Dataset}

\begin{figure}[htp!]
    \centering
    \includegraphics[width=0.6\linewidth]{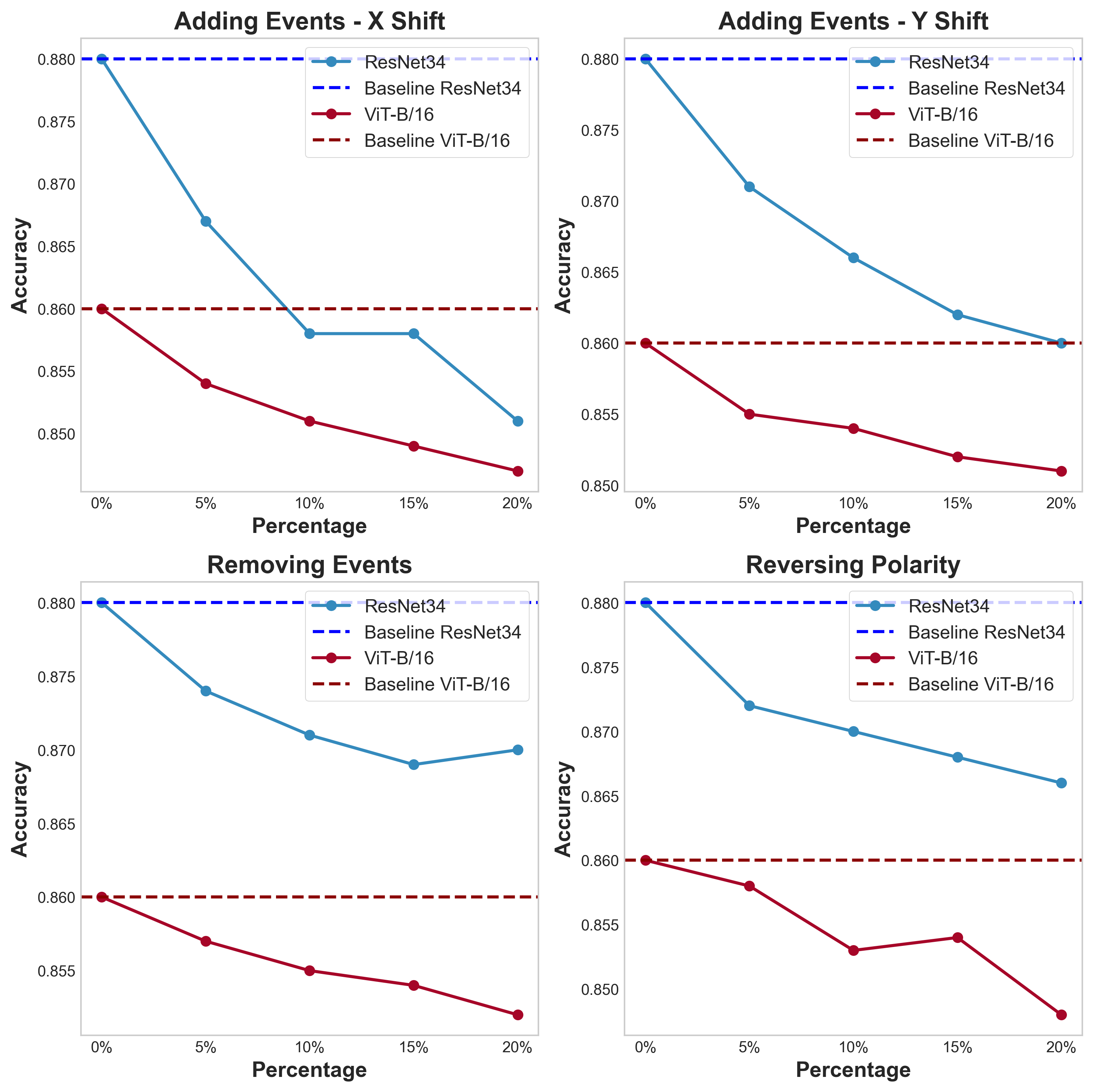}
    \caption{Performance of ResNet34 and ViT B16 across the simulated noise.}
    \label{fig:resnet_vit_noise_performance}
\end{figure}

The performance of the ResNet34 and ViT B16 models under varying noise conditions is shown in Figure \ref{fig:resnet_vit_noise_performance} and summarized in Table \ref{table:performance_noised_gen1}. As noise levels rise, both models experience a decrease in accuracy, although the patterns of this decline differ. Specifically, when events are shifted along the x and y axes, ResNet34's accuracy decreases more sharply than that of ViT B16. This trend is apparent in the X and Y shift plots, where ResNet34 shows a more pronounced drop in accuracy, particularly between 10\% and 20\% noise levels. This indicates that ResNet34 is more sensitive to event shifts, likely due to its reliance on spatial information, which is more directly affected by such transformations.

\begin{table}[htp!]
\centering
\caption{Performance of ResNet34 and ViT B16 on N-Caltech101, GEN1, and noisy GEN1 datasets. Accuracy is reported for different noise levels and types of noise.}
\includegraphics[width=\linewidth]{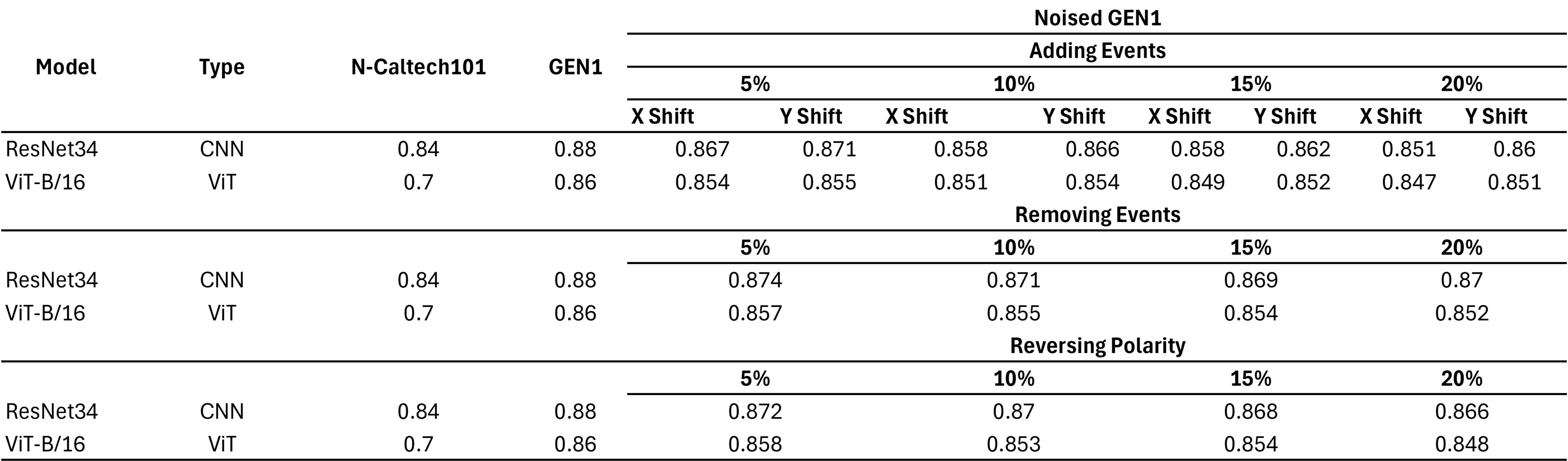}
\label{table:performance_noised_gen1}
\end{table}

\subsection{Performance Comparison: ResNet34 vs ViT B16}

When comparing ResNet34 and ViT B16 across clean and noisy datasets, ResNet34 holds a slight advantage in clean conditions, achieving higher accuracy and AUC values. However, ViT B16 surpasses ResNet34 in noisy environments, showing greater robustness to all noise types. This indicates that while ResNet34 is well-suited for controlled scenarios, ViT B16 offers better generalization in real-world, noisy applications, making it a more reliable choice for dynamic environments.

\subsection{Performance Comparison: ViT B16 vs State-of-the-art}

Compared to state-of-the-art models in event-based vision shown in Table \ref{table:calssification table}, ViT B16 showcases competitive performance. Although not tailored for event-based datasets, ViT B16's transformer architecture demonstrates strong adaptability and noise resilience. Its performance under noisy conditions, particularly for spatial shifts and polarity reversal, approaches or exceeds that of some specialized event-based models reported in the literature. This underscores the potential of transformer-based models in advancing event-based vision research, even without extensive pre-training on event-specific datasets.

\begin{table}[H]
    \centering
    \caption{State-of-the-art Classification Approaches on Event-based Vehicle Datasets.}
    \includegraphics[width=\linewidth]{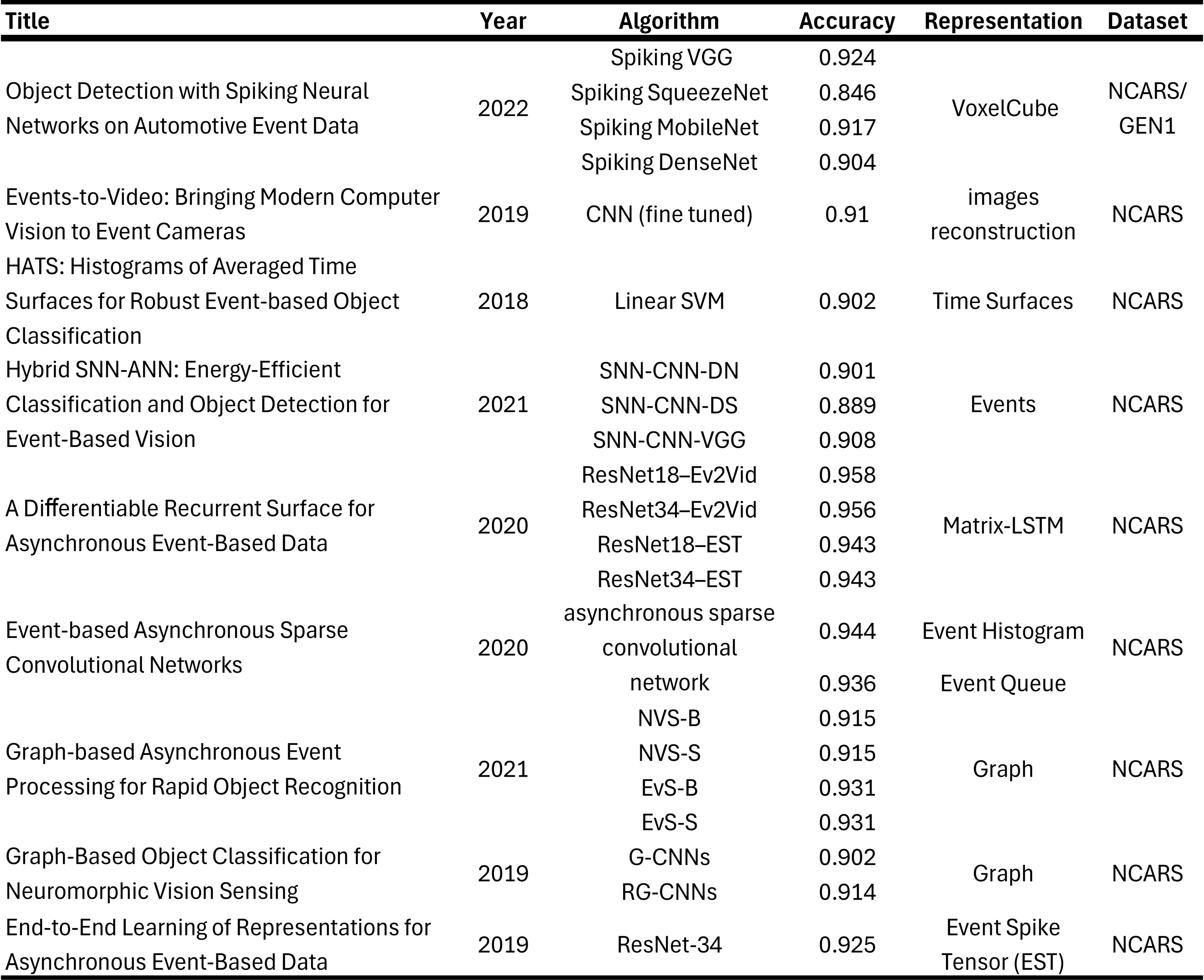}
    \label{table:calssification table}
\end{table}

\section{Conclusion}
\noindent In conclusion, this research fine-tuned ResNet34 and ViT B16 on the GEN1 event-based dataset to assess their performance in a classification task involving cars and pedestrians. Both models performed well, with ResNet34 achieving 88\% accuracy and ViT B16 following at 86\%. ViT B16 demonstrated strong potential, especially in classifying 76.6\% of cars and 90.3\% of pedestrians, despite being pre-trained on a smaller dataset. The analysis revealed that, while ResNet34 slightly outperformed ViT B16 in standard scenarios, ViT B16 showed greater resilience to noise, particularly with event shifts along the x and y axes. As noise levels increased, ViT B16's accuracy declined gradually, while ResNet34's dropped more sharply, suggesting that ViT B16 may be more robust to noisy data. This makes ViT B16 a promising candidate for real-world applications where noise is prevalent, such as UAVs and autonomous driving. \\

\noindent This study highlights the strengths and weaknesses of CNN-based and transformer-based models in noisy environments. The findings emphasize the potential of ViT models in applications where noise is inevitable, providing a foundation for future research to explore larger datasets or advanced noise-handling techniques to improve performance further. Additionally, this approach could be adapted for UAV applications, where noise and environmental factors can influence data quality, enhancing model reliability and performance in dynamic, real-world scenarios.

\bibliography{sample}

\end{document}